\documentclass[runningheads]{llncs}

\usepackage{eccv}

\usepackage{eccvabbrv}

\usepackage{graphicx}
\usepackage{booktabs}

\usepackage{amsmath}
\usepackage{amssymb}
\usepackage{multirow}
\usepackage{arydshln}
\usepackage{comment}
\usepackage{enumitem}
\usepackage{hhline}
\usepackage{multirow}

\usepackage[accsupp]{axessibility}  % Improves PDF readability for those with disabilities.

\usepackage{hyperref}

\usepackage{orcidlink}

\newcommand{\para}[1]{\noindent\textbf{#1}}

\begin{document}

\title{Learning Natural Consistency Representation for Face Forgery Video Detection} 

\titlerunning{NACO}

\author{Daichi Zhang\inst{1,2,3} \and
Zihao Xiao\inst{4} \and Shikun Li\inst{1,2} \and Fanzhao Lin\inst{1,2} \and Jianmin Li\inst{3} \and Shiming Ge\inst{1,2,}\thanks{Corresponding author}}

\authorrunning{D.~Zhang et al.}

\institute{
$^\text{1} $ Institute of Information Engineering, Chinese Academy of Sciences \\
$^\text{2} $ University of Chinese Academy of Sciences \\
$^\text{3 } $Department of Computer Science and Technology, Institute for AI, BNRist, Tsinghua University \quad
$^\text{4} $  RealAI\\
\email{\{zhangdaichi20,lishikun19,linfanzhao21\}@mails.ucas.ac.cn, lijianmin@mail.tsinghua.edu.cn, zihao.xiao@realai.ai, geshiming@iie.ac.cn}}

\maketitle

\begin{abstract}
Face Forgery videos have elicited critical social public concerns and various detectors have been proposed.
However, fully-supervised detectors may lead to easily overfitting to specific forgery methods or videos, and existing self-supervised detectors are strict on auxiliary tasks, such as requiring audio or multi-modalities, leading to limited generalization and robustness.
In this paper, we examine whether we can address this issue by leveraging visual-only real face videos.
To this end, we propose to learn the Natural Consistency representation (NACO) of real face videos in a self-supervised manner, which is inspired by the observation that fake videos struggle to maintain the natural spatiotemporal consistency even under unknown forgery methods and different perturbations.
Our NACO first extracts spatial features of each frame by CNNs then integrates them into Transformer to learn the long-range spatiotemporal representation, leveraging the advantages of CNNs and Transformer on local spatial receptive field and long-term memory respectively.
Furthermore, a Spatial Predictive Module~(SPM) and a Temporal Contrastive Module~(TCM) are introduced to enhance the natural consistency representation learning.
The SPM aims to predict random masked spatial features from spatiotemporal representation, and the TCM regularizes the latent distance of spatiotemporal representation by shuffling the natural order to disturb the consistency, which could both force our NACO more sensitive to the natural spatiotemporal consistency.
After the representation learning stage, a MLP head is fine-tuned to perform the usual forgery video classification task.
Extensive experiments show that our method outperforms other state-of-the-art competitors with impressive generalization and robustness.
\end{abstract}

\section{Introduction}
\label{sec:intro}

Recently, there has been a significant advancement in face forgery technology~\cite{lu2017attribute-guided,li2020advancing,li2023preim3d,thies2016face2face:}, particularly since the emergence of generative adversarial networks~(GANs)\cite{goodfellow2014generative}. The generated and manipulated faces are almost indistinguishable to the naked eye, but can be easily produced by available online tools like Deepfakes\footnote{https://github.com/deepfakes/faceswap} and FaceSwap\footnote{https://github.com/MarekKowalski/FaceSwap}. This accessibility enables perpetrators easily using these techniques to generate forgery videos to mislead the public, defame celebrities, or even fabricate evidence, which could result in severe social, political, and security threats~\cite{DBLP:journals/tog/SuwajanakornSK17}. Hence, how to develop effective face forgery video detectors is crucial to prevent the malicious applications of these techniques.

Various detectors have been proposed to address the face forgery video detection task.
Existing supervised detectors mainly focuses on specific patterns which are discriminative in forgery and real videos, such as frequency domain information~\cite{DBLP:conf/eccv/QianYSCS20}, specific artifacts~\cite{DBLP:conf/cvpr/WangW0OE20,DBLP:conf/cvpr/FeiDYSX022,DBLP:conf/cvpr/LiL19c}, and spatiotemporal clues~\cite{DBLP:conf/iccv/ZhengB0ZW21,DBLP:conf/ijcai/ZhangLL0G21,nguyen2019multi,DBLP:conf/wifs/AfcharNYE18,DBLP:conf/ijcai/HuXWLW021,wang2023altfreezing,li2018in,yang2019exposing}.
However, these fully-supervised detectors may easily overfit to specific forgery methods or videos in training datasets and the artifacts they rely on may be corrupted under perturbations, leading to limited generalization and robustness. Recent FTCN~\cite{DBLP:conf/iccv/ZhengB0ZW21} explores temporal coherence by setting spatial kernel size to one and AltFreezing~\cite{wang2023altfreezing} proposes to learn spatial and temporal features respectively to improve the generalization. But FTCN ignores all the clues from spatial dimension and AltFreezing neglects the connection between spatial and temporal domains, which may hinder their performance.
Other self-supervised detectors aim to explore model-agnostic self-supervised representation to detect, such as LipForensics~\cite{DBLP:conf/cvpr/HaliassosVPP21} pretrains on lipreading dataset focusing on mouth movements and the RealForensics~\cite{DBLP:conf/cvpr/HaliassosMPP22} pretrains on visual-audio modalities to avoid overfitting. However, LipForensics requires labeled datasets for pretraining and only focuses on the mouth region, RealForensics requires both visual-audio modalities for pretraining and performs visual multi-task during finetuning, which are strict on the auxiliary tasks and the pretaining datasets that may limit their scalability and performance.

\begin{figure}[t]
    \centering
    \includegraphics[width=.9\linewidth]{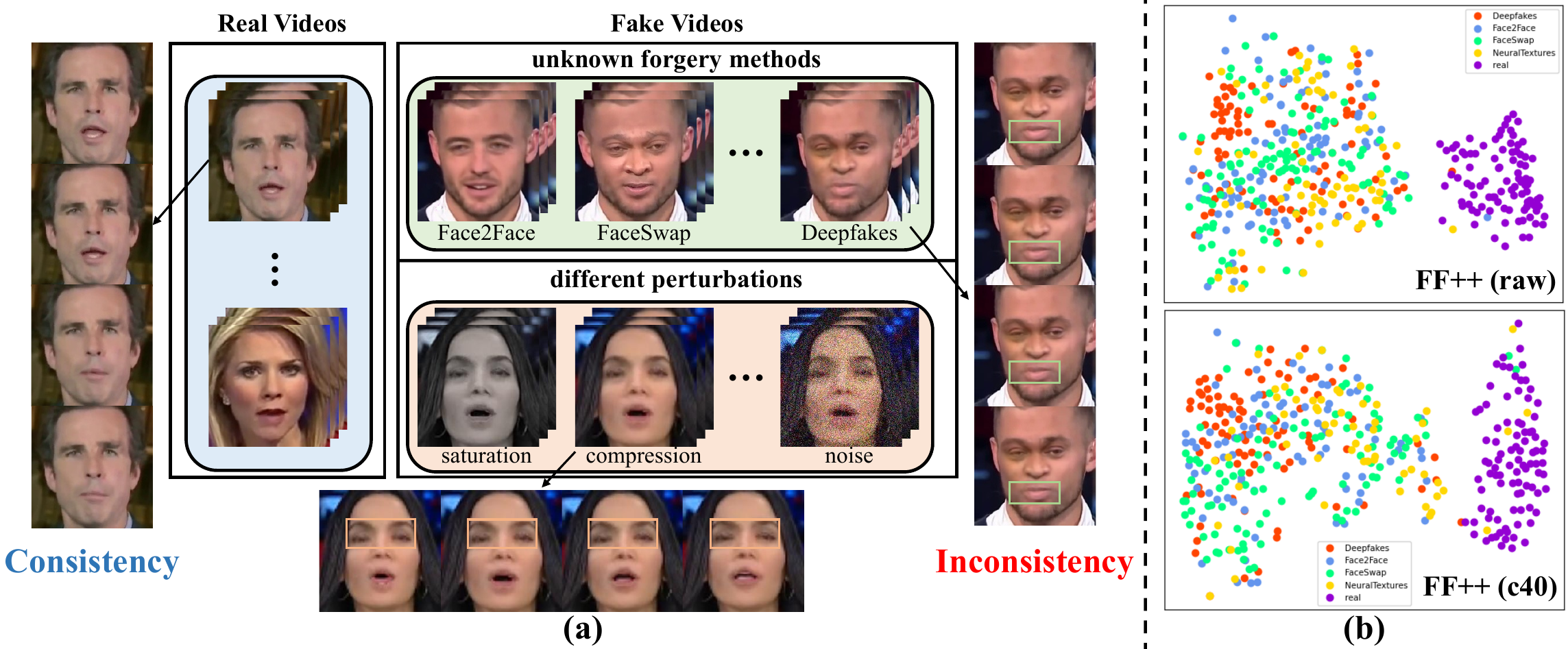}
    \caption{(a) Real face videos exhibit natural spatiotemporal consistency while fake videos generated from unknown forgery methods or under different perturbations both show inconsistencies. (b) t-SNE~\cite{tSNE} visualization of our NACO on uncompressed~(raw) and heavily compressed~(c40) FF++ which includes four different forgery methods.
    }\label{figure:mot}
\end{figure}

With all the concerns above in mind, we raise the question: whether we can learn a general and robust representation on visual-only real videos in a self-supervised manner by fully leveraging their spatiotemporal clues, without requiring any fake videos, thus avoiding overfitting to specific forgery methods or videos.
To achieve this, we are inspired by the observation that fake videos struggle to preserve the natural spatiotemporal consistency in real face videos, defined as the semantic-level spatiotemporal coherence in natural face videos~(the opposite as inconsistency), such as facial movements and expression changes. This high-level natural consistency is corrupted even under unknown forgery methods and different perturbations, providing the insight to leverage this as clue to improve generalization and robustness. As shown in Fig.~\ref{figure:mot} (a), the real videos exhibit natural spatiotemporal consistency while the fake videos generated by unknown forgery methods or under different perturbations both show inconsistency.
If we could leverage this natural consistency of real face videos, we could embed all real face videos into one compact cluster in latent space while all other fake videos~(from unknown generation methods or different perturbations) are embedded into another cluster with a clear discrepancy margin, such as shown in Fig.~\ref{figure:mot} (b), achieving general and robust face forgery video detection.

To this end, we propose to learn the Natural Consistency~(NACO) representation of visual-only real face videos in a self-supervised manner.
We first model the spatiotemporal representation of videos by initially extracting spatial features from each frame by CNNs then integrating the spatial feature sequence into Transformer to learn the long-term spatiotemporal representation. 
The intuition behind this design is we assume the natural clues of real videos exist in both single frame and long-range sequence. And due to the low information density~\cite{DBLP:journals/corr/abs-2111-06377}, pixel space could not provide crucial information for natural consistency representation learning. But CNNs and Transformer have their inherent advantages in local spatial receptive field and long-term memory respectively~\cite{DBLP:conf/cvpr/HeZRS16,DBLP:conf/iclr/DosovitskiyB0WZ21,ramachandran2019stand,wang2018non}. By incorporating both, we could explore the spatiotemporal representation from local spatial and long-range dimensions to enhance the representation learning.

Further, two specifically designed self-supervised tasks, Spatial Predictive Module~(SPM) and Temporal Contrastive Module~(TCM), are introduced to enhance the natural consistency learning. The SPM aims to reconstruct random masked spatial features from learned spatiotemporal representation by incorporating a CNN decoder. The TCM regularizes the latent distance of spatiotemporal representation pairs by shuffling the original frame order, which disturbs the natural consistency. The intuition for these two modules is that we assume a desired NACO representation could use the spatiotemporal context to predict the missing information for SPM and should be sensitive to the frame order which indicates the natural consistency for TCM.
{Leveraging the self-supervised natural consistency learning, our detector does not so easily overfit the forgery methods or videos in training datasets as in supervised methods.}
Finally, a MLP head is fine-tuned guided by NACO representation to perform the usual forgery video classification task.

In brief, our contributions are summarized as follows:
(1) We propose to learn the Natural Consistency representation~(NACO) of visual-only real face videos for general and robust face forgery detection, which leverages the advantages of CNNs and Transformer on local spatial receptive field for single frame and long-term memory for frame sequence respectively.
(2) Two specifically designed self-supervised tasks are introduced, including Spatial Predictive Module~(SPM) and Temporal Contrastive Module~(TCM), which both serve to enhance the natural consistency learning on real face videos.
(3) Extensive experiments on public datasets demonstrate the superiority of our proposed method over the state-of-the-art competitors with impressive generalization and robustness.

\section{Related Work}
\label{sec:related}

\para{Face Forgery Video Detection.}
Since high-fidelity face forgery videos cause severe threats to society, how to detect them has become an urgent and essential issue. 
Recent deep-learning based detectors have achieved impressive performance for both fully-supervised and self-supervised detectors.
Existing fully-supervised detectors naively train a supervised binary classifier based on specific patterns to distinguish between real and fake videos, such as frequency~\cite{DBLP:conf/eccv/QianYSCS20}, artifacts~\cite{DBLP:conf/cvpr/WangW0OE20,DBLP:conf/cvpr/FeiDYSX022,DBLP:conf/cvpr/LiL19c,hua2021interpretable}, and spatiotemporal clues~\cite{DBLP:conf/iccv/ZhengB0ZW21,DBLP:conf/ijcai/ZhangLL0G21,nguyen2019multi,DBLP:conf/wifs/AfcharNYE18,DBLP:conf/ijcai/HuXWLW021,wang2023altfreezing,li2018in,yang2019exposing}.
However, these specific patterns can be easily corrupted under common perturbations or eliminated by other unknown forgery methods, leading to limited generalization and robustness. 
Other self-supervised detectors aim to leverage model-agnostic representation to detect, such as LipForensics~\cite{DBLP:conf/cvpr/HaliassosVPP21} and RealForensics~\cite{DBLP:conf/cvpr/HaliassosMPP22}. However, LipForensics requires labeled datasets for pretraining and only focuses on the mouth region, RealForensics requires both visual-audio modalities for pretraining and performs visual multi-task during finetuning, which are strict on the auxiliary tasks and the pertaining datasets that may limit their scalability and performance.
In contrast, we aim to leverage the natural consistency of visual-only real face videos without additional requirements in both local receptive spatial and long-range spatiotemporal dimensions to learn the model-agnostic representation for general and robust detection.

\para{Self-Supervised Learning.}
The most common strategy in self-supervised learning is designing auxiliary tasks to introduce supervision without labels~\cite{DBLP:journals/corr/abs-2111-06377,xie2022simmim,zhang2016colorful,noroozi2016unsupervised,gidaris2018unsupervised}.
Contrastive learning, which pulls positive pairs closer and pushes negative pairs away in latent space, has achieved impressive performance in self-supervised representation learning~\cite{oord2018representation,chen2020simple,wang2015unsupervised,li2022selective,zhang2022leverage}.
Masked models also demonstrate impressive representation capacity by masking part of the input and forcing the model to predict them by leveraging the context~\cite{DBLP:journals/corr/abs-2111-06377,xie2022simmim,li2020look,tong2022videomae}. 
Some self-supervised detectors focus on specific pattern clues~\cite{DBLP:journals/corr/abs-2203-12208,li2020face,DBLP:journals/corr/abs-2204-08376,DBLP:conf/nips/ChenZSWL22,DBLP:journals/tomccap/GeLLZ0022,zhang2023self}, which are susceptible to perturbations and unknown forgery methods.
Others require large labeled or multi-modal datasets for pretraining~\cite{DBLP:journals/corr/abs-2203-01265,DBLP:conf/cvpr/HaliassosVPP21,DBLP:conf/cvpr/HaliassosMPP22,DBLP:journals/corr/abs-2301-01767}, which are strict on the auxiliary tasks and the pertaining datasets that may limit their generalization and scalability. 
Different from them, we focus on the natural consistency of visual-only real videos without additional requirements for both pretraining and finetuning phases with two specifically designed auxiliary tasks~(SPM and TCM) in both spatiotemporal dimensions.

\para{Vision Transformer.}
Transformers have achieved impressive performance in various natural language processing tasks by introducing self-attention mechanism~\cite{DBLP:conf/nips/VaswaniSPUJGKP17}. Inspired by this, researchers in computer vision field also seek to explore its potential applications. %with multilayer perceptron
The vision transformer~(ViT)~\cite{DBLP:conf/iclr/DosovitskiyB0WZ21} could be the first to apply transformer to computer vision tasks by treating image patches as token sequence. 
Since then, various ViT-based works have been proposed for different vision tasks, such as semantic segmentation~\cite{duke2021sstvos,DBLP:conf/iccv/StrudelPLS21} and object detection~\cite{DBLP:conf/iclr/DosovitskiyB0WZ21}. 
ViT has also been applied to face forgery detection task~\cite{DBLP:journals/corr/abs-2207-06612,dong2022protecting,DBLP:conf/iccv/ZhengB0ZW21,DBLP:journals/corr/abs-2203-01265,DBLP:journals/corr/abs-2301-01767}.
However, these methods either directly adopt ViT to train an end-to-end supervised classifier while ignoring its self-supervised representation capacity, especially on long-range memory~\cite{dong2022protecting,DBLP:conf/iccv/ZhengB0ZW21,DBLP:journals/corr/abs-2207-06612}, or train in a self-supervised way on multi-modalities~\cite{DBLP:journals/corr/abs-2203-01265,DBLP:journals/corr/abs-2301-01767}. 
In contrast, our method incorporates Transformer with CNNs to achieve visual-only natural consistency representation learning in both local spatial receptive~\cite{DBLP:conf/cvpr/HeZRS16,ge2018low,ge2017detecting} and long-term spatiotemporal dimensions.

\section{Method}
\label{sec:method}

In this section, we discuss our proposed NACO which consists of two stages, as illustrated in Fig.~\ref{figure:pipeline}. The first stage aims to learn the NACO representation by first modeling the spatiotemporal representation of videos and then employing two designed self-supervised tasks~(SPM and TCM) for natural consistency learning on visual-only real face videos. 
The NACO representations are then used to guide the binary forgery video classification task in the second stage by optimizing a MLP head. The details of each stage are presented below.

\subsection{Spatiotemporal Representation}
\label{subsec:framework}

We aim to develop a general and robust face forgery video detector by leveraging the natural consistency of real face videos as clues. 
Hence, the initial step is to model the spatiotemporal representation of video. 
We first question whether the representation can be learned directly from pixel space. Due to its low information density~\cite{DBLP:journals/corr/abs-2111-06377}, the pixel space lacks crucial information for natural consistency learning.
Further, based on our observations, we intuitively assume that the natural clues in real videos exist in both individual frames and long sequences. Thus, our method aims to model the spatiotemporal representation from two distinct aspects: local receptive spatial (single frame) and long-range spatiotemporal (long-range sequence) dimensions, described in detail as follows:

\begin{figure*}[ht]
    \centering
    \includegraphics[width=\linewidth]{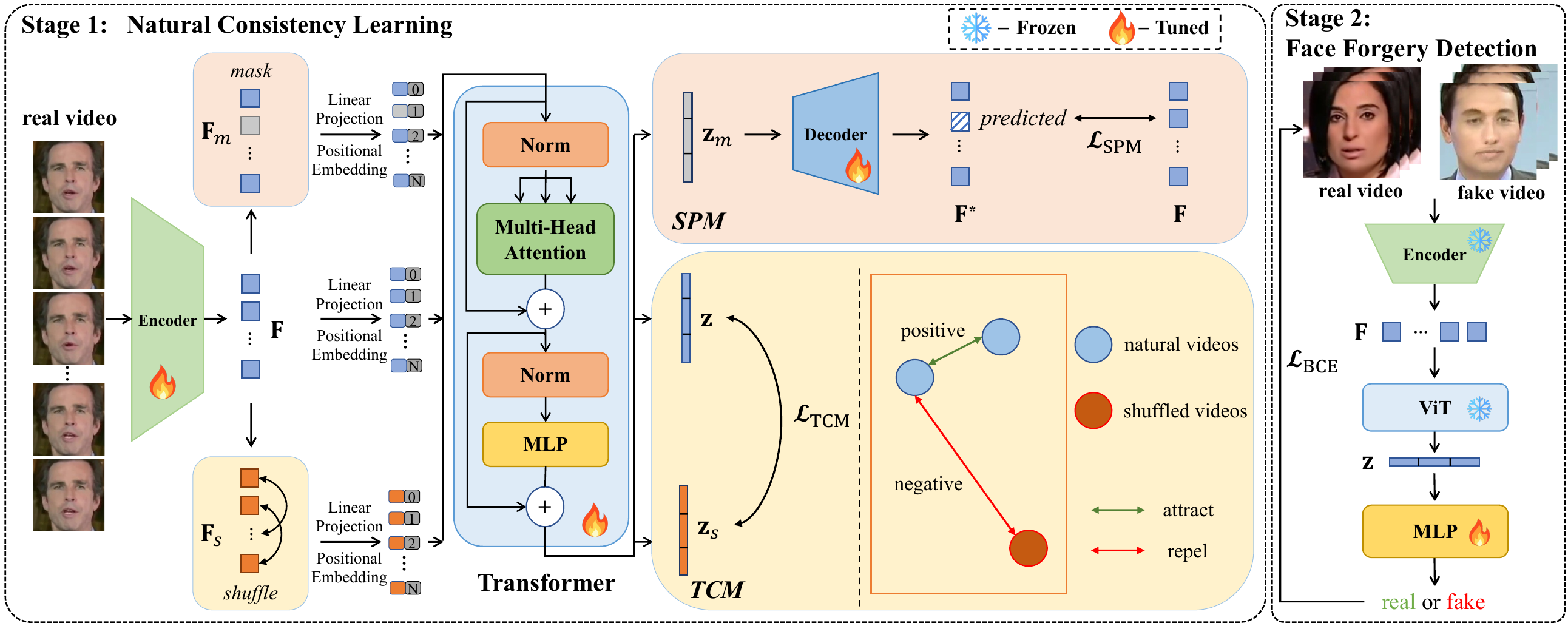}
    \caption{\label{figure:pipeline} The pipeline of our proposed NACO. 
    In the first stage, real videos are first extracted into spatial feature sequence~$\mathbf{F}$ by CNN encoder, which is fed into the Transformer to learn long-range spatiotemporal representation~$\mathbf{z}$. Further, two designed auxiliary tasks: SPM and TCM are introduced to enhance the natural consistency learning on real face videos.
    In the second stage, the encoder and Transformer are frozen and a fully-connected classification head~(two-layer MLP) guided by learned NACO representation is optimized to perform the usual face forgery video classification task.}
\end{figure*}

\para{Local Receptive Spatial Feature.} 
Instead of learning directly from pixel space, we initially introduce a simply designed CNNs to extract the local receptive spatial feature of each frame, which we assume can provide more local spatial natural clues by CNN's inherent advantage on local spatial receptive field~\cite{DBLP:conf/cvpr/HeZRS16}.
We assume access to a real face video dataset~$\mathcal{D}_r$. Each sample in $\mathcal{D}_r$ represents a real face video consisting of multiple frames. We sample random clips~$\mathbf{X}$ from each video with consistent $n$ frames employing face extraction and alignment to get the input frame sequence of our model, where~$\mathbf{X} = \left\{ \mathbf{x}_i\right\}_{i=1}^{n}$ and $\mathbf{X} \in \mathcal{D}_r$.
Then we introduce a simple CNN encoder to extract the spatial feature of each frame~$\mathbf{x}_i$, which projects the frame sequence into spatial feature sequence $\mathbf{F}$, formulated as follows:
\begin{equation}
    \mathbf{F} = \operatorname{Conv}(\mathbf{X},\Theta_e) = \operatorname{Conv}(\left\{ \mathbf{x}_i\right\}_{i=1}^{n}, \Theta_e) = \left\{ \mathbf{f}_i\right\}_{i=1}^{n},
\end{equation}
where the~$\{\Theta_e\}$ is the parameters of encoder and the dimension of~$\mathbf{f}_i$ is 256.

\para{Long-range Spatiotemporal Representation.} 
We further explore the natural clues in the long-range sequence to learn the long-term spatiotemporal representation of real face videos. To achieve this, we incorporate Transformer for its inherent advantage in long-term memory~\cite{DBLP:conf/iclr/DosovitskiyB0WZ21,ramachandran2019stand,wang2018non} by integrating the extracted spatial features.
Specifically, we reshape $\mathbf{f}_i$ into $16 \times 16$ tokens and employ a trainable linear projection matrix~$\mathbf{W}$ with positional embedding~$\mathbf{E}_{pos}$ to get the input feature sequence and feed them into our Transformer to learn the spatiotemporal representation~$\mathbf{z}$ of each video clip, which mainly consists of $K$ standard Transformer Encoder blocks~\cite{DBLP:conf/nips/VaswaniSPUJGKP17}, each block contains a multi-head self-attention~(MSA) block and an MLP block with the commonly-used LayerNorm~(LN) before them. We also use the GELU as the activation function and the whole process can be formulated as follows:
\begin{align}
    & \mathbf{z}_0 = \mathbf{W} \cdot \mathbf{F} + \mathbf{E}_{pos} = \mathbf{W} \cdot \left[ \mathbf{f}_1, \cdots, \mathbf{f}_n\right]^{\mathrm{T}} + \mathbf{E}_{pos},\\
    & \mathbf{z}_k = \operatorname{MSA}(\operatorname{LN}(\mathbf{z}_{k-1})) + \mathbf{z}_{k-1}, \qquad k=1\ldots K
\end{align}
We denote all the trainable parameters of the Transformer as~$\{\Theta_v\}$ and regard the last-layer output~$\mathbf{z}_K$ as the learned spatiotemporal representation with dimension of 768.
%, where~$\operatorname{dim}(\mathbf{z}_K) = 768$.
Unless stated otherwise, the mentioned representation~$\mathbf{z}$ in the following represents the representation learned from the last-layer~$\mathbf{z}_K$, which we denote as $\mathbf{z}$ for simplicity.
Furthermore, two designed self-supervised tasks are introduced to enhance the natural consistency representation learning in latent space, described in detail in the following.

\subsection{Natural Consistency Learning}
\label{subsec:stage1}
We assume that low-level clues such as artifacts could be corrupted by unknown manipulation types or different perturbations, which causes limited generalization and robustness. But the high-level clues, such as the natural spatiotemporal consistency would still exist under both situations, since forgery methods typically struggle to preserve such clues during generation and high-level features are naturally less susceptible to common perturbations.
Hence, we aim to leverage the natural consistency of visual-only real videos in a self-supervised manner to develop a general detector that can achieve both high generalization and robustness. 
Based on the spatiotemporal representation learned from visual-only real videos described above, we further design two self-supervised tasks: Spatial Predictive Module~(SPM) and Temporal Contrastive Module~(TCM) to enhance natural consistency learning, described in detail below. 

\para{Spatial Predictive Module~(SPM).}
We assume that an effective natural consistency representation should learn the relationship between consistent spatial features at different timestamps and could use the context to predict the missing information, preventing from relying on specific input. Some previous works have also demonstrated this~\cite{xie2022simmim,DBLP:journals/corr/abs-2111-06377}. Thus, we initially randomly mask parts of extracted local spatial features~$\mathbf{f}_i$ before feeding them into Transformer. Then we introduce a CNN decoder to predict the masked spatial features.

To make this process clearer, we first define one operation between two sequence:~$\mathbf{A} \ominus \mathbf{B}$ which indicates the indices at which the element is not masked in~$\mathbf{A}$ but masked in~$\mathbf{B}$. For example, if~$\mathbf{A} = \{\mathbf{a}\ \mathbf{b}\ \mathbf{c}\ \mathbf{d}\}$ and~$\mathbf{B} = \{\mathbf{a}\ [\operatorname{mask}]\ \mathbf{c}\ [\operatorname{mask}]\ \}$, then we have~$\mathbf{A} \ominus \mathbf{B} = \{2,4\}$.

Then for extracted local spatial feature sequence~$\mathbf{F}$, we first random mask parts of~$\mathbf{F}$ with a ratio~$\alpha$ to get the masked local spatial feature sequence~$\mathbf{F}_m = \operatorname{Mask}(\mathbf{F}, \alpha)$, where there are total~$n\times\alpha$ spatial features~$\mathbf{f}_i$ are randomly masked by setting the tensor value to zeros. Then the~$\mathbf{F}_m$ is fed into Transformer to learn the spatiotemporal representation~$\mathbf{z}_m$ of the masked sequence.

Further, a CNN decoder is introduced to predict the masked local spatial feature from the learned masked spatiotemporal representation~$\mathbf{z}_m$. We calculate the distance between the original and predicted sequence to regularize the prediction process. Specifically, we only compute the distance on masked spatial features, which can be formulated as follows:

\begin{eqnarray}
\label{eq:spm}
    \mathcal{L}_\textsubscript{SPM}&=&\left\| {\mathbf{F}^{*}} - \mathbf{F} \right\| \nonumber    \\
    ~&=&\left\| {\mathrm{Conv}(\mathbf{z}_{m}, \Theta_d)} - \mathbf{F} \right\| \nonumber   \\
    ~&=&\sum_{i \in \mathbf{F} \ominus \mathbf{F}_m}\left\| \mathbf{f}_{i}^{*} - \mathbf{f}_{i} \right\|,
\end{eqnarray}
where~$\mathbf{F}^*$ is the predicted local spatial feature sequence and~$\{\Theta_d\}$ is the parameters of the CNN decoder. We use the mean squared error~(MSE loss) as the distance function.

\para{Temporal Contrastive Module~(TCM).}
We further hypothesize that a desired natural consistency representation should be sensitive to the frame sequence order, which also indicates the natural consistency. 
To address this, we aim to regularize the distance of spatiotemporal representation in latent space using contrastive learning by disturbing the original frame order. We initially disturb the natural consistency in real face videos by random shuffling the frame sequence order, which is also equivalent to shuffling the extracted local spatial features order to get~$\mathbf{F}_s = \operatorname{Shuffle}(\mathbf{F})$, where~$\mathbf{F}_s \neq \mathbf{F}$.

Then we introduce contrastive learning to regularize the distance in latent space. The key question here is how to construct the positive and negative pairs for contrastive learning.
We first define the similarity of two different spatiotemporal representation~$\left( \mathbf{z}_i, \mathbf{z}_j \right)$ learned by the Transformer, which can be formulated as follows:
\begin{equation}
    \label{eq:sim}
    \operatorname{sim} \left(\mathbf{z}_i, \mathbf{z}_j\right) = \frac{\mathbf{z}_i \cdot \mathbf{z}_j}{\mathop{\max}(\left\| \mathbf{z}_i \right\|_{2} \cdot \left\| \mathbf{z}_j \right\|_{2}, \epsilon)},
\end{equation}
where hyper-parameter~$\epsilon$ is set to $1e-8$.
Then for a natural unshuffled local spatial feature sequence~$\mathbf{F}$ and corresponding spatiotemporal representation~$\mathbf{z}$, we consider the representation learned from other natural unshuffled input as positive pairs~$\mathbf{z}^+$, while the~$\mathbf{z}_s$ from shuffled spatial feature sequence~$\mathbf{F}_s$ as negative pairs~$\mathbf{z}^-$.
Thus, we can formulate the contrastive loss~$\mathcal{L}_\textsubscript{TCM}$ within mini-batch samples as follows:
\begin{small}
\begin{equation}
    \label{eq:tcm}
    \mathcal{L}_\textsubscript{TCM}=-\log \frac{\exp \left( \operatorname{sim}\left(\mathbf{z}, \mathbf{z}^+ \right) / \tau\right)}{\exp \left(\operatorname{sim} \left(\mathbf{z}, \mathbf{z}^+ \right) / \tau\right)+\sum \exp \left( \operatorname{sim} \left(\mathbf{z}, {\mathbf{z}^-} \right) / \tau\right)},
\end{equation}
\end{small}
where the temperature $\tau$ is set to 0.5. Specifically, for each input video clip during each iteration, we shuffle its order to get one negative sample.

Therefore, the total loss function for the self-supervised natural consistency representation learning stage can be formulated as follows:
\begin{gather} 
    \label{eq:ssl}
    \mathcal{L} = \lambda_1 \mathcal{L}_\textsubscript{SPM} + \lambda_2 \mathcal{L}_\textsubscript{TCM},
\end{gather}
where $\{\lambda_1, \lambda_2\}$ are the hyper-parameter weights to balance the SPM and TCM loss.
Optimizing by these two designed tasks, we finally obtain the desired Natural Consistency representation~(NACO) by $\mathbf{z}^* = f(\mathbf{z}, \Theta_e, \Theta_v, \Theta_d)$, where $f(\cdot)$ represents the natural consistency learning process.

\subsection{Face Forgery Detection}
\label{subsec:stage2}

After the natural consistency representation learning, we freeze the backbone including the encoder and Transformer, 
and add a fully-connected classification head~(two-layer MLP) which takes the learned NACO representation to classify real and fake videos. 
Since fake videos can't preserve the natural consistency as real videos, they should be discriminated by our NACO generally.
Given an input video~$\mathbf{\widehat{X}}$, the detection can be formulated as follows:
\begin{equation}
    \label{eq:mlp}
    y = \operatorname{MLP}(\mathbf{\hat{z}^*}, \Theta_c) = \operatorname{MLP}(g(\mathbf{\widehat{X}}, \Theta_e, \Theta_v), \Theta_c),
\end{equation}
where the $\left\{ \Theta_e, \Theta_v, \Theta_c \right\}$ denote the parameters of the encoder, Transformer, and classification head, and~$g(\cdot)$ means the process to obtain NACO representation from $\left\{ \Theta_e, \Theta_v \right\}$ during fine-tuning. %Noticing that $\left\{ \Theta_e, \Theta_v \right\}$ are frozen and 
Noticing that only $\left\{ \Theta_c \right\}$ is optimized in this stage while others are frozen.
Then given a labeled forgery video dataset which includes both real and fake videos,%~$\mathcal{D} = \left\{ \left( {x}_i, {y}_i \right) \right\}_{i=1}^{N}$, 
we choose the vanilla binary cross-entropy loss~$\mathcal{L}_\textsubscript{BCE}$ to supervise the forgery classification task.

\section{Experiment}
\label{sec:exp}

\subsection{Experimental Settings}

\para{Datasets.}
We choose the large-scale real face video dataset: VoxCeleb2~\cite{DBLP:conf/interspeech/ChungNZ18} for pretraining, which contains over 1 million utterances from 6,112 celebrities, extracted from videos on YouTube.
For the face forgery video detection task, we choose following datasets for finetuning and evaluation: 
(1) FaceForensics++(FF++)~\cite{Rssler2019FaceForensics} contains 1,000 real videos and 4,000 fake videos with three different compression levels~(raw/c23/c40, from uncompressed to heavily compressed) generated from four different manipulation methods, including two face swapping methods, Deepfakes~\footnote{https://github.com/deepfakes/faceswap}~(DF) and FaceSwap~\footnote{https://github.com/MarekKowalski/FaceSwap}~(FS), and two face reenactment methods, Face2Face~\cite{thies2016face2face:}~(F2F) and NeuralTextures~\cite{thies2019deferred}~(NT). Unless stated otherwise, we use the mildly compressed version of the dataset~(c23).
(2) Celeb-DF-v2(CDF)~\cite{DBLP:conf/cvpr/LiYSQL20} is a challenging dataset including 590 real videos and 5,639 fake videos.
(3) DFDC~\cite{dolhansky2019the} is a subset of the Deepfake Detection Challenge Dataset~\footnote{https://www.kaggle.com/c/deepfake-detection-challenge/data}, where each video is recorded in challenging environments.
(4) FaceShifter(FSh)~\cite{li2020advancing} is the recent high-fidelity face swapping method that has been applied to the real videos of FF++.
(5) DeeperForensics(DFo)~\cite{DBLP:conf/cvpr/JiangLW0L20} contains real videos recorded in difficult real-world scenarios and high-fidelity forgery videos based on the real videos from FF++.

\para{Evaluation metrics.}
Following recent works~\cite{DBLP:conf/cvpr/HaliassosVPP21,DBLP:conf/iccv/ZhengB0ZW21,DBLP:conf/cvpr/HaliassosMPP22,wang2023altfreezing}, we report the video-level Area Under the Receiver Operating Characteristic Curve~(AUC(\%)) and Accuracy~(ACC(\%)) to compare with prior works. For frame-level detectors, the metrics are averaged over sampled video frames. 
%Please refer to our supplementary material for more details about the datasets and baselines.

\para{Implementation details.}
For each video, we sample continuous 20 frames reshaped into 224$\times$224 from a random offset as input sequence and repeat three times to get the mean results, which should be beneficial for different video lengths.
Then we employ face extraction and alignment using tool~\footnote{https://github.com/1adrianb/face-alignment}. 
The encoder consists of three convolutional layers with 3$\times$3 kernel and 64, 128, 256 channels and one average pooling layer which extract spatial features of 256 dimensions, and the decoder is one convolutional layer with 1$\times$3 kernel which projects the spatiotemporal representation of 768 dimensions into 256 spatial features. We choose the basic ViT-Base architecture described in~\cite{DBLP:conf/iclr/DosovitskiyB0WZ21} with $K=12$ blocks, whose final output is 768 dimensions representation. We use a two-layer MLP including one hidden layer of 256 dimensions and one output layer with Softmax activation as our classification head.
We employ the Adam optimizer with an initial learning rate of~$5e-4$ and a weight decay of~$1e-4$. 
The batch size is set to 64 and we empirically set~$\alpha=1/2$ in SPM, $\lambda_1 = 1.0$ and $\lambda_2 = 0.5$ in Eq.~(\ref{eq:ssl}).

\subsection{Experimental Results}
\label{sec:exp_result}

\begin{table}[ht]
\small
\centering
\caption{\textbf{Generalization to unseen datasets.} Video-level AUC~(\%) is reported. Other methods' results are from~\cite{DBLP:conf/cvpr/HaliassosMPP22} and their original papers.}
\resizebox{0.65\columnwidth}{!}{
\begin{tabular}{lccccc}
\hline
Method       & CDF & DFDC & FSh & DFo & \textit{Avg} \\
\hline
Xception \cite{Rssler2019FaceForensics} & 73.7 & 70.9 & 72.0 & 84.5 & 75.3\\
CNN-aug \cite{DBLP:conf/cvpr/WangW0OE20} & 75.6 & 72.1 & 65.7 & 74.4 & 72.0\\
Patch-based \cite{DBLP:conf/eccv/ChaiBLI20} & 69.6 & 65.6 & 57.8 & 81.8 & 68.7\\
Two-branch \cite{DBLP:conf/eccv/MasiKMGA20} & 76.7 & - & - & - & -\\
Face X-ray \cite{li2020face} & 79.5 & 65.5 & 92.8 & 86.8 & 81.2\\ 
Multi-task \cite{nguyen2019multi}      & 75.5  & 68.1 & 66.0 & 77.7 & 71.9\\
DSP-FWA \cite{DBLP:conf/cvpr/LiL19c}             & 69.5  & 67.3 & 65.5 & 50.2 & 63.1\\
CNN-GRU \cite{sabir2019recurrent}      & 69.8  & 68.9 & 80.8 & 74.1 & 73.4\\
LipForensics-scratch~\cite{DBLP:conf/cvpr/HaliassosVPP21}   & {62.5} & {65.5} & {84.7} & {84.8} & 74.4\\ 
LipForensics \cite{DBLP:conf/cvpr/HaliassosVPP21}   & 82.4  & 73.5 & 97.1 & 97.6 & 87.7\\
FTCN \cite{DBLP:conf/iccv/ZhengB0ZW21}      & \underline{86.9}  & {74.0} & {98.8} & 98.8 & 89.6\\ 
RealForensics-scratch \cite{DBLP:conf/cvpr/HaliassosMPP22}  & {69.4} & {68.1} & {87.9} & {89.3} & 78.7\\
RealForensics \cite{DBLP:conf/cvpr/HaliassosMPP22}     & \underline{86.9} & \underline{75.9} & \textbf{99.7} & \underline{99.3} & \underline{90.5}\\
ISTVT \cite{zhao2023istvt}                              & {84.1} & {74.2} & {99.3} & {98.6} & {89.1}\\
NoiseDF \cite{wang2023noise}                            & {75.9} & {63.9} & {-} & {70.9} & {-}\\
AltFreezing \cite{wang2023altfreezing}                  & \textbf{89.5} & {-} & \underline{99.4} & \underline{99.3} & {-}\\
\hline
NACO-scratch~(ours)                & {67.9} & {69.4} & {88.3} & {89.5} & {78.8}\\
NACO~(ours)           & \textbf{89.5}  & \textbf{76.7} & \underline{99.4} & \textbf{99.5} & \textbf{91.2}\\
%\bottomrule
\hline
\end{tabular}}
\label{table:cross-dataset}
\end{table}

\para{Generalization to unseen datasets.}
We first evaluate our method's generalization by training on FF++ then evaluating on other four unseen challenging datasets as presented in Tab.~\ref{table:cross-dataset} (scratch means directly fine-tuning without pretraining).
We observe that both frame-level detectors, such as~\cite{Rssler2019FaceForensics,DBLP:conf/cvpr/WangW0OE20} and simply designed supervised video-level detectors~\cite{sabir2019recurrent} lead to limited generalization. But the video-level detectors focusing on spatiotemporal features achieve better performance, such as~\cite{DBLP:conf/cvpr/HaliassosVPP21,DBLP:conf/iccv/ZhengB0ZW21,wang2023altfreezing,DBLP:conf/cvpr/HaliassosMPP22} and our NACO, providing additional evidence that high-level spatiotemporal clues are the key for generalization.

Furthermore, our NACO outperforms other state-of-the-art supervised~\cite{DBLP:conf/iccv/ZhengB0ZW21,wang2023altfreezing} and self-supervised video-level detectors~\cite{DBLP:conf/iccv/ZhengB0ZW21,wang2023altfreezing}
with the highest average 91.2\% AUC score across the four unseen datasets, indicating the superiority by leveraging natural consistency of real face videos. And the performance is also better when training from scratch compared to LipForensics and RealForensics.
The promising results suggest the potential of our method to detect more challenging unseen forgery videos in the future.

\begin{table}[ht]
\small
\centering
\caption{\textbf{Generalization to unseen manipulations.} Video-level AUC~(\%) on each subset of FF++ is reported. Other methods' results are from~\cite{DBLP:conf/cvpr/HaliassosMPP22} and original papers.}
\resizebox{.65\columnwidth}{!}{
\begin{tabular}{lccccc}
\hline
\multirow{2}{*}{Method} & \multicolumn{4}{c}{Train on remaining three} & \multirow{2}{*}{\textit{Avg}}     \\ \cline{2-5}
 & DF & FS & F2F & NT &  \\ \hline
Xception~\cite{Rssler2019FaceForensics}             & 93.9 & 51.2 & 86.8 & 79.7 & 77.9 \\
CNN-aug~\cite{DBLP:conf/cvpr/WangW0OE20}            & 87.5 & 56.3 & 80.1 & 67.8 & 72.9 \\
Patch-based\cite{DBLP:conf/eccv/ChaiBLI20}          & 94.0 & 60.5 & 87.3 & 84.8 & 81.7 \\
Face X-ray~\cite{li2020face}                        & 99.5 & 93.2 & 94.5 & 92.5 & 94.9 \\
%\hline
CNN-GRU~\cite{sabir2019recurrent}                   & 97.6 & 47.6 & 85.8 & 86.6 & 79.4 \\
LipForensics-scratch~\cite{DBLP:conf/cvpr/HaliassosVPP21}   & {93.0} & {56.7} & {98.8} & {98.3} & {86.7} \\ 
LipForensics~\cite{DBLP:conf/cvpr/HaliassosVPP21}   & {99.7} & 90.1 & \underline{99.7} & {99.1} & {97.1} \\ 
FTCN \cite{DBLP:conf/iccv/ZhengB0ZW21}             & \underline{99.9} & \textbf{99.9} & \underline{99.7} & \underline{99.2} & \textbf{99.7}\\
RealForensics-scratch \cite{DBLP:conf/cvpr/HaliassosMPP22}  & {98.8} & {87.9} & {98.7} & {88.6} & {93.5}\\
RealForensics \cite{DBLP:conf/cvpr/HaliassosMPP22}  & \textbf{100.} & {97.1} & \underline{99.7} & \underline{99.2} & \underline{99.0}\\
AltFreezing \cite{wang2023altfreezing}  & {99.8} & \underline{99.7} & {98.6} & {96.2} & \underline{98.6}\\
\hline
NACO-scratch (ours)       & {98.9} & {88.3} & {97.9} & {89.1} & {93.6}\\
NACO~(ours)                                          & \underline{99.9} & \underline{99.7} & \textbf{99.8} & \textbf{99.4} & \textbf{99.7}\\ \hline
\end{tabular}}
\label{tab:cross-manipulation} 
\end{table}

\para{Generalization to unseen manipulations.}
We further evaluate our method's generalization to unknown manipulation methods on each subset of FF++ by training on three and evaluating on the remaining one as shown in Tab.~\ref{tab:cross-manipulation}.
We observe that both frame-level detectors that focus on low-level artifacts~\cite{Rssler2019FaceForensics} and naive supervised video-level detectors~\cite{sabir2019recurrent} suffer significant drops when detecting unknown forgery types. 
But the detectors which leverage high-level spatiotemporal clues achieve better performance, such as~\cite{DBLP:conf/cvpr/HaliassosVPP21,DBLP:conf/iccv/ZhengB0ZW21,wang2023altfreezing,DBLP:conf/cvpr/HaliassosMPP22} and our NACO. This also supports our motivation that the inconsistency still exists in unknown forgery types, regardless of face swapping or reenactment.
Moreover, our method outperforms other state-of-the-art supervised~\cite{DBLP:conf/iccv/ZhengB0ZW21,wang2023altfreezing} and self-supervised video-level methods~\cite{DBLP:conf/cvpr/HaliassosVPP21,DBLP:conf/cvpr/HaliassosMPP22} with achieving 99.7\% average AUC, indicating the effectiveness of leveraging natural consistency representation in visual-only real face videos to defend unseen manipulation methods. 
Besides, the generalization is also better when training from scratch compared to~\cite{DBLP:conf/cvpr/HaliassosVPP21,DBLP:conf/cvpr/HaliassosMPP22}.

Furthermore, we also compare the parameters and architectures in Tab.~\ref{tab:params-arch}. We observe that our method which only optimizes a MLP head achieves the minimum finetuning parameters with impressive generalization.

\begin{table}[ht]
\small
   \centering
   \caption{\textbf{Trainable parameters for usual face forgery detection task and generalization comparisons.} Video-level AUC~(\%) is reported when trained on FF++. 2D/3D means the CNN architectures and TF means the Transformer.
   }\label{tab:params-arch}
   \resizebox{.65\linewidth}{!}{
       \begin{tabular}{lccccc}
          %\toprule
          \hline
              {Method} & {\#params} & Arch & FSh & DFo \\
              \hline
              {LipForensics-scratch} \cite{DBLP:conf/cvpr/HaliassosVPP21} & 36.0M & 2D+MS-TCN & 84.7 & 84.8\\ 
              {LipForensics} \cite{DBLP:conf/cvpr/HaliassosVPP21} & 24.8M & 2D+MS-TCN & 97.1 & 97.6\\
              {FTCN} \cite{DBLP:conf/iccv/ZhengB0ZW21} & 26.6M & 3D+TF & 98.8 & 98.8\\
              {RealForensics-scratch} \cite{DBLP:conf/cvpr/HaliassosMPP22} & 21.4M & 2D+CSN & {87.9} & {89.3}\\
              {RealForensics} \cite{DBLP:conf/cvpr/HaliassosMPP22} & 21.4M & 2D+CSN & \textbf{99.7} & \underline{99.3}\\
              AltFreezing \cite{wang2023altfreezing} & 27.2M & 3D & \underline{99.4} & \underline{99.3}\\
              \hline
              {NACO~(ours)} & \textbf{4.7M} & 2D+TF & \underline{99.4} & \textbf{99.5}\\
          %\bottomrule
          \hline
       \end{tabular}
   }
\end{table}

\begin{table*}[ht]
\small
   \centering
   \caption{\textbf{Robustness to unseen perturbations.} Video-level AUC~(\%) on FF++ under seven different perturbations described in~\cite{DBLP:conf/cvpr/JiangLW0L20}.
   Other methods' results are from~\cite{DBLP:conf/cvpr/HaliassosMPP22} and $^*$ denotes our reproduction.
   }\label{tab:robust}
   \resizebox{.9\linewidth}{!}{
       \begin{tabular}{lccccccccl}
          %\toprule
          \hline
              {Method} & %\textcolor{gray}{Origin} & 
              {Clean} & {Saturation} & {Contrast} & {Block} & {Noise} & {Blur} & {Pixel} & {Compress} & \textit{Avg/Drop}\\
              \hline
              {Xception~\cite{Rssler2019FaceForensics}} & \textcolor{gray}{99.8} & 
              99.3 & 98.6 & \textbf{99.7} & 53.8 & 60.2 & 74.2 & 62.1 & 78.3/-21.5\\
              {CNN-aug~\cite{DBLP:conf/cvpr/WangW0OE20}} & \textcolor{gray}{99.8} & 
              99.3 & 99.1 & 95.2 & 54.7 & 76.5 & 91.2 & 72.5 & 84.1/-15.7\\
              {Patch-based~\cite{DBLP:conf/eccv/ChaiBLI20}} & \textcolor{gray}{99.9} & 
              84.3 & 74.2 & 99.2 & 50.0 & 54.4 & 56.7 & 53.4 & 67.5/-32.4\\
              {X-Ray~\cite{li2020face}} & \textcolor{gray}{99.8} & 
              97.6 & 88.5 & 99.1 & 49.8 & 63.8 & 88.6 & 55.2 & 77.5/-22.3\\
              %\hline
              {CNN-GRU~\cite{sabir2019recurrent}} & \textcolor{gray}{99.9} & 
              99.0 & 98.8 & 97.9 & 47.9 & 71.5 & 86.5 & 74.5 & 82.3/-17.6\\
              {LipForensics} \cite{DBLP:conf/cvpr/HaliassosVPP21} & \textcolor{gray}{99.9} & \textbf{99.9} & {99.6} & 87.4 & 73.8 & 96.1 & 95.6 & 95.6 & 92.6/-7.3\\
              {FTCN} \cite{DBLP:conf/iccv/ZhengB0ZW21} & \textcolor{gray}{99.4} & 
              99.4 & 96.7 & 97.1 & 53.1 & 95.8 & {98.2} & 86.4 & 89.5/-9.9\\
              {RealForensics} \cite{DBLP:conf/cvpr/HaliassosMPP22} & \textcolor{gray}{99.8} & 
              99.8 & {99.6} & 98.9 & 79.7 & 95.3 & {98.4} & \textbf{97.6} & 95.6/-4.2\\
              AltFreezing$^*$ \cite{wang2023altfreezing} & \textcolor{gray}{99.9} & {99.5} & \textbf{99.8} & {97.1} & {75.2} & \textbf{97.4} & {98.1} & {92.6} & {94.2/-5.7}\\
              \hline
              {NACO~(ours)} & \textcolor{gray}{99.9} & 98.9 & 98.2 & 98.4 & \textbf{86.5} & {96.2} & \textbf{98.6} & {96.7} & \textbf{96.2}/\textbf{-3.7}\\
          %\bottomrule
          \hline
       \end{tabular}}
   %}
\end{table*}

\para{Robustness to unseen perturbations.}
We further investigate our model's robustness against various unseen perturbations by training on uncompressed videos and evaluating on videos added perturbations. 
We consider the following seven perturbations described in~\cite{DBLP:conf/cvpr/JiangLW0L20}: saturation, contrast, block-wise, Gaussian noise, Gaussian blur, pixelation, and video compression. %(H.$264$ codec). 
Each perturbation is applied under five different severity levels.
We report the average video-level AUC scores across all severity levels and make comparisons in Tab.~\ref{tab:robust}.
We find our NACO suffers significantly less than frame-level detectors, such as~\cite{li2020face,Rssler2019FaceForensics}. This indicates leveraging high-level natural consistency leads to more robust detection than relying on low-level clues corrupted in perturbations, which is consistent with our initial motivation.
Furthermore, our method also outperforms other recent supervised and self-supervised video-level SOTAs~\cite{DBLP:conf/cvpr/HaliassosVPP21,DBLP:conf/iccv/ZhengB0ZW21,DBLP:conf/cvpr/HaliassosMPP22,wang2023altfreezing}, with the minimum AUC drop of -3.7\%, especially on Gaussian noise, blur, and pixelation, which disturb more on consistency. This also provides additional evidence of our method's superior robustness to unseen perturbations by leveraging the natural consistency of real videos.

\subsection{Ablation Study}
\label{sec:ablation}
We conduct further ablations by fine-tuning on FF++~(c23) and testing on FF++~(c40) and Celeb-DF~(CDF) to evaluate the robustness and generalization. 
%Please refer to our supplementary material for more additional experimental results.

\begin{figure}[ht]
   \centering
   \includegraphics[width=.75\linewidth]{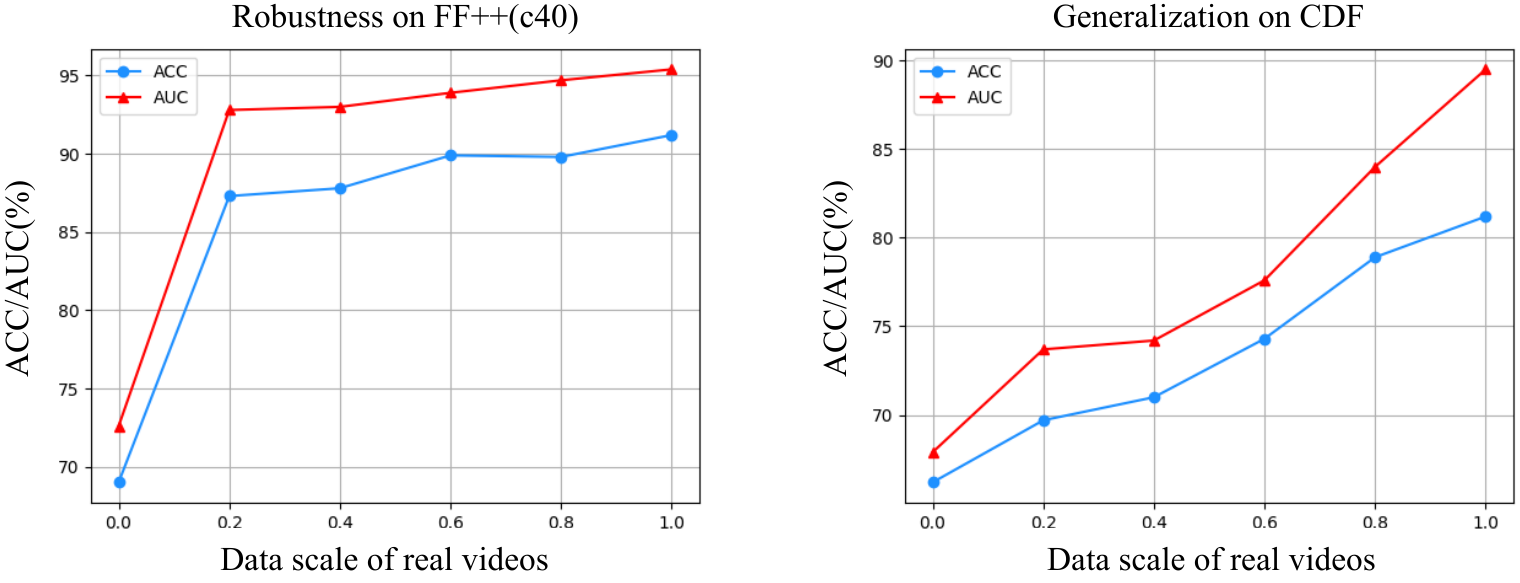}
   \caption{
   \textbf{Comparisons on different number of real samples} in natural consistency representation learning stage.
   }\label{figure:data-scale}
\end{figure}

\para{Number of real samples.}
We first investigate how the quantity of real face videos used for natural consistency learning affects our model's performance. 
We regard the whole original number of real samples as 1.0 and reduce the scale with an interval of 0.2, noticing that setting the data scale to 0.0 means we directly fine-tune the model with random initialization without pretraining stage.
The results are presented in Fig.~\ref{figure:data-scale}.
We can see that our method benefits from a large number of real samples,
which proves the effectiveness of our proposed NACO by leveraging the natural consistency representation on visual-only real face videos.

\para{Natural Consistency Learning.}
We further investigate how the two designed self-supervised tasks~(SPM and TCM) for natural consistency learning effect our model's performance by employing each respectively as presented in Tab.~\ref{tab:ablation-loss}. 
We observe that employing both modules achieve the highest performance on both robustness and generalization, average 1.05\% and 9.60\% AUC improvements, which indicates both designed tasks have positive effect on the natural consistency representation learning on real face videos. 

\begin{table}[ht]
\small
   \centering
   \caption{
   \textbf{Analysis of the two designed self-supervised tasks}~(SPM and TCM) for natural consistency learning.
    }\label{tab:ablation-loss}
   \resizebox{.6\linewidth}{!}{
       \begin{tabular}{cccccc}
          %\toprule
          \hline
              \multirow{2}{*}{$\mathcal{L}_\textsubscript{SPM}$} & \multirow{2}{*}{$\mathcal{L}_\textsubscript{TCM}$} &
              \multicolumn{2}{c}{FF++~(c40)} &
              \multicolumn{2}{c}{CDF} \\
              \hhline{~~|--||--|}
              & & {ACC~(\%)} & {AUC~(\%)} & {ACC~(\%)} & {AUC~(\%)}\\
              \hline
              $\checkmark$ & {-} & {90.7} & {94.6} & {77.8} & {80.6} \\
              {-} & $\checkmark$ & {87.8} & {94.1} & {74.3} & {79.2} \\
              $\checkmark$ & $\checkmark$ & {91.2} & {95.4} & {81.2} & {89.5} \\
          %\bottomrule
          \hline
       \end{tabular}
    }
\end{table}

\subsection{Forgery Localization}
\begin{figure}[ht]
   \centering
   \includegraphics[width=.75\linewidth]{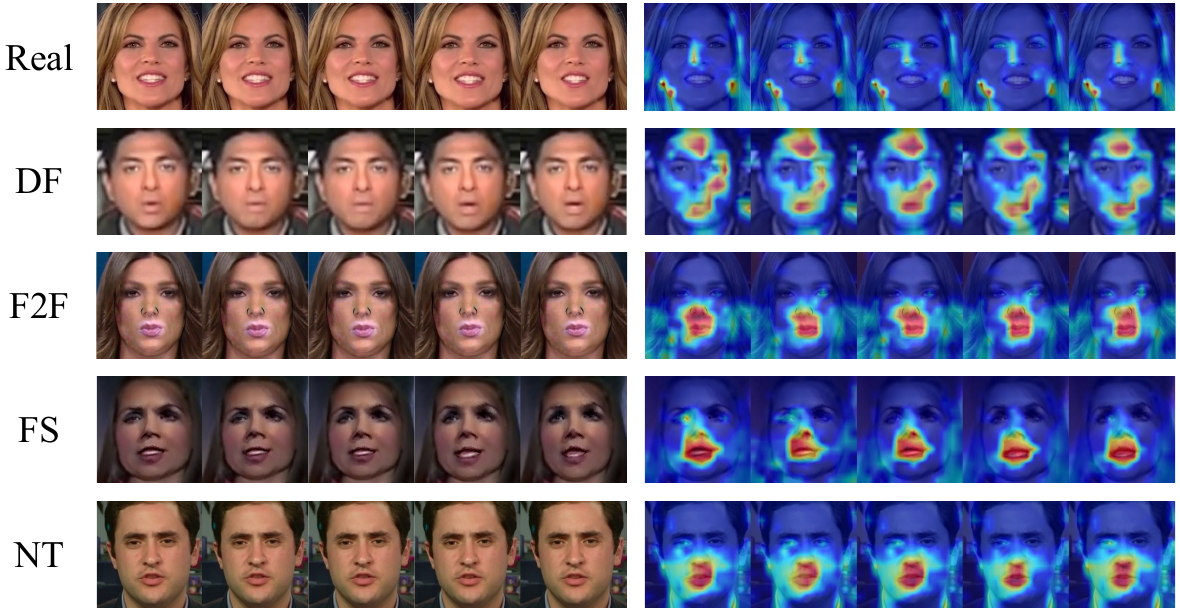}
   \caption{\textbf{Forgery localization.} Grad-CAM results on five consecutive frames on FF++~(c23). 
   We find that our method can effectively respond to the inconsistencies in fake videos and localize the forgery areas.
   }\label{figure:grad-cam}
\end{figure}

Furthermore, we provide the Grad-CAM~\cite{DBLP:conf/iccv/SelvarajuCDVPB17} response based on the predicted spatial features on five consecutive frames from each subset of FF++~(c23) in Fig.~\ref{figure:grad-cam}.
From the results, we observe that the responses of forgery videos focus on the specific face areas, such as the mouth and center face, where exists inconsistency. But the response of real videos is average over the entire facial area since there are no inconsistencies in natural real videos. 
The results indicate that our method can effectively localize the inconsistent forgery areas in fake videos and provide human-trustable explanations for decision results.

\section{Conclusion}
\label{sec:conclusion}

In this paper, we propose to learn the Natural Consistency representation~(NACO) of visual-only real face videos to develop a general and robust face forgery detector. 
NACO initially extracts spatial features of each single frame by CNNs then integrates them into Transformer to learn long-range spatiotemporal representation.
Furthermore, two specifically designed self-supervised tasks, Spatial Predictive Module~(SPM) and Temporal Contrastive Module~(TCM), are introduced to enhance the natural consistency learning on visual-only real face videos.
Extensive experiments have shown that our method achieves impressive generalization to unknown forgery types and robustness to various perturbations.
In the future, we aim to apply our model to more complicated samples from real scenarios, such as from the web, and also extend our method to other media forensic tasks, such as audio and multi-modalities.

%\clearpage  % TODO FINAL: This \clearpage needs to be removed from both review and camera-ready versions.

\section*{Acknowledgements}
This work was partially supported by grants from the Pioneer R\&D Program of Zhejiang Province (2024C01024), Open Research Project of the State Key Laboratory of Media Convergence and Communication, Communication University of China (SKLMCC2022KF004).

\bibliographystyle{splncs04}
\bibliography{main}
\end{document}